%% file: iclr2027_conference.tex
\documentclass{article} 
\usepackage{iclr2027_conference,times}

\input{math_commands.tex}

\usepackage{hyperref}
\usepackage{url}
\usepackage{booktabs}
\usepackage{multirow}
\usepackage{graphicx}
\usepackage{xcolor}
\usepackage{colortbl}
\usepackage{wrapfig}

\title{Can Visual Generation Expand What Unified Multimodal Models Know?}

\usepackage[auth-lg]{authblk} %

\author{Wenkai Li\thanks{Equal contribution} \,}
\author{Wenyi Hong\protect\footnotemark[1] \,}
\author{Zhuoyi Yang\protect\footnotemark[1] \,}
\author{Jie Tang\thanks{Corresponding author}}
\affil{Tsinghua University \quad Z.AI}
\affil{\texttt{lwk21@mails.tsinghua.edu.cn, wenyi.hong@outlook.com, zhuoyiyang2000@gmail.com, jietang@tsinghua.edu.cn}}

\makeatletter
\newcommand{\blfootnote}[1]{%
  \begingroup
  \renewcommand{\thefootnote}{}%
  \footnotetext{#1}%
  \endgroup
}
\makeatother

\iclrfinalcopy 
\renewcommand{\headrulewidth}{0pt} 
\begin{document}

\maketitle

\fancyhead{} 
\blfootnote{Work was done when WL, WH, ZY interned at Z.AI. Project page: \url{https://kevinlllee.github.io/Knowledge-Transfer-Probing/}}

\begin{figure}[h]
\begin{center}
\includegraphics[width=\linewidth]{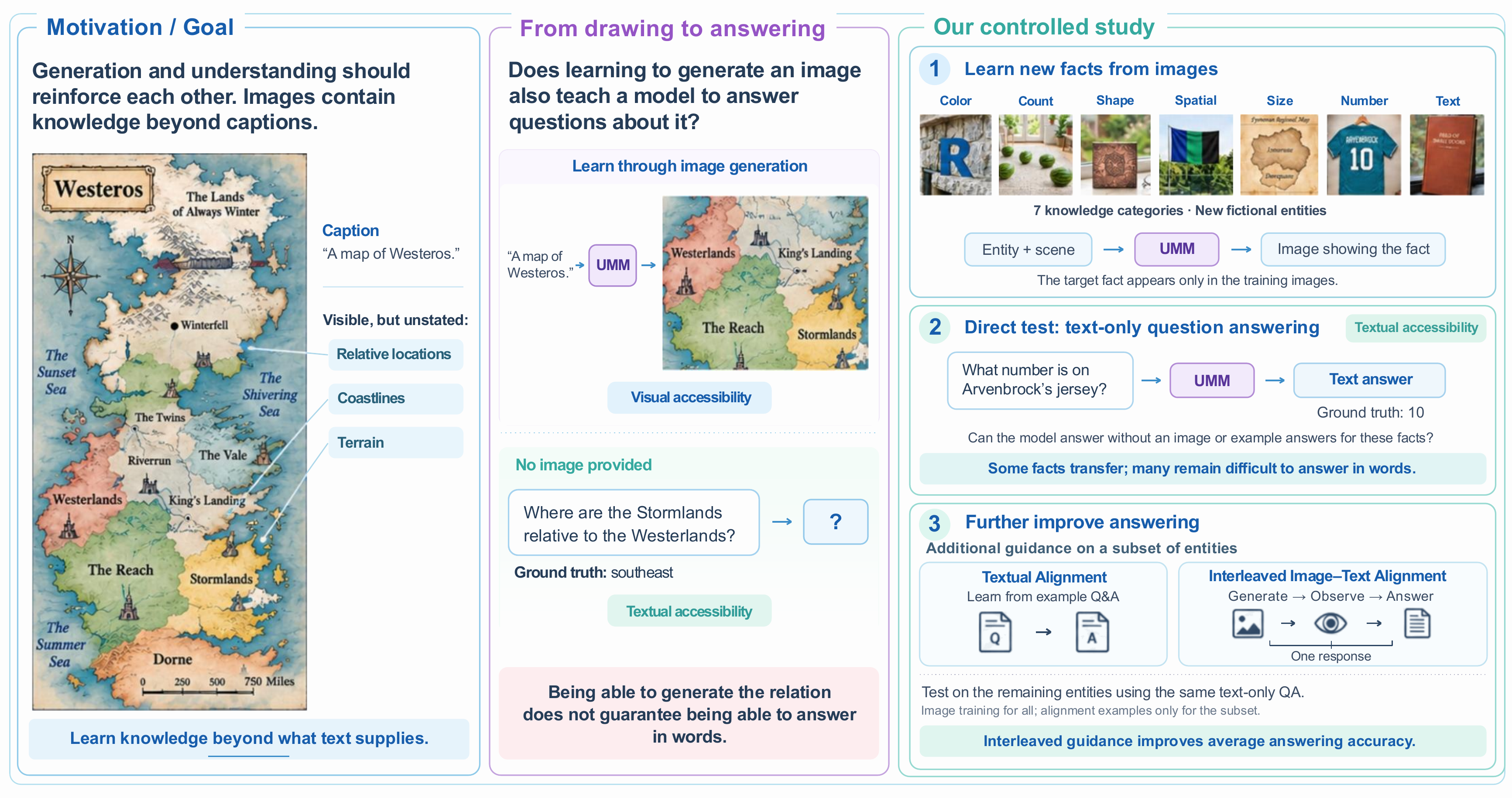}
\end{center}
\vspace{-2mm}
\caption{
\textbf{Can visual generation expand what unified multimodal models know?}
\textbf{Left:} Images contain rich visual information that accompanying text may leave unspecified.
\textbf{Middle:} Knowledge reproduced in generated images is not necessarily accessible through text-only QA.
\textbf{Right:} We first supply new facts only through training images and test text-only QA across seven knowledge categories.
Direct transfer is partial; alignment on a subset of entities can improve access on the remaining entities, with interleaved image--text alignment yielding higher average accuracy than textual alignment.
}
\label{fig:overview}
\end{figure}

\begin{abstract}
Unified multimodal models (UMMs) receive image-generation supervision that exposes them to visual information not fully documented in text.
Can learning to generate images expand what pretrained models know?
We introduce Knowledge Transfer Probing (KTP), a controlled protocol that supplies new knowledge through image-generation training and tests its accessibility through text-only question answering.
Across seven knowledge categories and three UMMs, we observe direct but uneven transfer from visual supervision to language.
Motivated by these observations, we successively introduce Textual Alignment and Interleaved Image--Text Alignment, which connect generation and answering through shared knowledge and interleaved responses, respectively.
Guidance on selected entities can improve knowledge access on other entities without QA supervision, with additional gains from interleaved alignment.
Further analysis shows that knowledge correctly reproduced in images is often more accessible through text, although substantial gaps remain between the two forms of access.
These findings support visual generation as a source of knowledge beyond textual supervision, and motivate joint investigation of training data design and model architecture to make this knowledge more broadly accessible.
The dataset and evaluation protocol is available at: \url{https://github.com/kevinlllee/Knowledge-Transfer-Probing}.
\end{abstract}

\input{section/1_intro}
\input{section/2_related_work}
\input{section/3_1_exp_setup}
\input{section/3_2_exp1_i2t_transfer}
\input{section/3_3_exp2_text_alignment}

\input{section/3_4_exp3_interleaved_alignment}
\input{section/3_5_exp4_ablation_loss_mask}
\input{section/3_6_exp5_infer_visual_cot}
\input{section/3_7_exp6_visual_textual_association}
\input{section/5_conclusion}

\subsection*{AI use statement}

We used AI tools to assist in implementing the data construction pipeline and used model APIs to generate data in batches.
For experimental analysis, AI tools helped develop small analysis scripts, including those for computing and comparing conditional probabilities, and supported discussions of research ideas and experimental findings.
During manuscript preparation, we used AI tools for translation and language editing.
All AI-assisted writing was manually reviewed by the authors.

\subsection*{Reproducibility statement}

In Appendix~\ref{app:data_construction}, we describe the data construction pipeline, including image generation, editing, quality control, and the creation of captions and alignment data.
In Appendix~\ref{app:training_details}, we summarize the data sampling ratios and training hyperparameters.
Finally, in Appendix~\ref{app:evaluation_protocol}, we specify the criteria for evaluating textual and visual accessibility, including answer matching and edit-distance thresholds.

\bibliography{iclr2027_conference}
\bibliographystyle{iclr2027_conference}

\appendix

\input{appendix/training_details}

\end{document}

%% file: math_commands.tex
\usepackage{amsmath,amsfonts,bm}

\def\eqref#1{equation~\ref{#1}}

\def\1{\bm{1}}

\DeclareMathAlphabet{\mathsfit}{\encodingdefault}{\sfdefault}{m}{sl}
\SetMathAlphabet{\mathsfit}{bold}{\encodingdefault}{\sfdefault}{bx}{n}



%% file: section/1_intro.tex
\section{Introduction}
\label{sec:intro}

A key aspiration of unified multimodal models (UMMs) is to enable generation and understanding to reinforce one another,  enabling more capable foundation models~\citep{synergy2026}.
This aspiration is particularly compelling because the visual world contains information that language does not fully document~\citep{tong2026beyond}.
A photograph of a local building or traditional artifact may reveal its structure, ornamentation, and spatial relationships, even when the accompanying caption mentions only its name or location. 
By learning to generate images, a model receives supervision over visual content that need not be explicitly described in the accompanying text.
Whether this learning also benefits understanding, however, remains an important question: \textit{can learning to generate images expand what unified models know?}

Knowledge sharing does not follow automatically from combining generation and understanding within one model.
UMMs such as Janus-Pro and BAGEL report strong performance on separate visual understanding and generation benchmarks~\citep{januspro2025, bagel2025}.
\citet{tong2026beyond} further demonstrates task-level benefits of multimodal co-training.
These results, however, do not establish whether specific knowledge learned through generation becomes usable in language.
\textit{Modal Aphasia} highlights this distinction: unified models can reproduce learned visual concepts while struggling to describe them in words~\citep{modal2026}.
In this work, we study the \emph{language accessibility} of visually acquired knowledge: 
The extent to which visual knowledge can be retrieved and correctly accessed and used in language.

We develop Knowledge Transfer Probing (KTP), a controlled protocol across seven visual knowledge categories, using fictional entity--attribute associations to reduce interference from pretrained knowledge.
During text-to-image (T2I) training, prompts identify entities and describe scenes, while target attribute values appear only in output images.
We test transfer through text-to-text (T2T) QA that asks for
an entity's attribute value without an input image.
We experiment across UMMs with different architectures: BAGEL, SenseNova-U1 \citep{sensenova2026}, and GLM-Omni.
The results show that T2I training alone enables correct textual answers for some of this knowledge, demonstrating direct transfer from visual supervision to language.
The extent of transfer varies across models and knowledge categories.
Much of the knowledge correctly expressed in generated images remains absent from these initial answers.
These results establish the potential of visual generation as a source of knowledge.

We further investigate whether auxiliary supervision during training can improve knowledge transfer.
All entities receive T2I supervision, while a subset also receives alignment examples.
We compare two forms of training supervision: \emph{Textual Alignment} provides text-only QA examples, whereas \emph{Interleaved Image--Text Alignment} combines image generation, visual observation, and answering in a single response.
The evaluation is conducted using the same T2T QA protocol on the remaining entities, whose target attribute values appear only in the T2I training images.
Both forms of alignment improve textual answers for these entities, with interleaved alignment yielding larger gains.
These results show that auxiliary supervision can improve access to knowledge beyond directly taught answers.
The advantage of interleaved alignment further demonstrates that the organization of training supervision matters for knowledge transfer.
Figure~\ref{fig:overview} summarizes our motivation and main findings.
We will release the dataset and evaluation protocol.

Our contributions are threefold:

\begin{itemize}
    \item \textbf{We introduce Knowledge Transfer Probing, a controlled protocol to test whether visual generation expands what UMMs know.}
    The protocol controls the source of target knowledge and reduces interference from pretrained associations.
    It evaluates whether knowledge supplied through generation-training images becomes usable in text-only QA, taking a first step toward measuring the knowledge contribution of visual generation.

    \item \textbf{We demonstrate direct but incomplete knowledge transfer from visual generation to language.}
    Across UMMs of varying architectures, generation supervision enables text-only answers for some knowledge, while much of the knowledge expressed in generated images remains absent from initial answers.
    Transfer varies substantially across models and knowledge categories.

    \item \textbf{We show that targeted supervision improves knowledge access beyond directly supervised entities.}
    Interleaved image--text alignment yields larger gains than text-only alignment, establishing supervision design as an effective means of strengthening transfer.
    Together with the observed cross-model differences, these findings motivate joint investigation of architecture and data organization for knowledge sharing in UMMs.
\end{itemize}

%% file: section/2_related_work.tex
\section{Related Work}

\textbf{Unified Multimodal Models.} UMMs combine visual understanding and image generation. Chameleon and Emu3 autoregressively model discrete multimodal tokens \citep{chameleon2024,emu32024}, while Transfusion, Show-o, and Show-o2 combine autoregressive text modeling with diffusion or flow matching for visual generation \citep{transfusion2024,showo2025,showo22025}. Janus-Pro, BAGEL, and SenseNova-U1 scale training and explore different designs for coupling these capabilities \citep{januspro2025,bagel2025,sensenova2026}. Controlled studies highlight visual-space alignment and architectural specialization as conditions for cross-task generalization and understanding-generation synergy \citep{generalization2026,synergy2026}.

\textbf{Cross-Modal Knowledge Transfer.} Cross-modal knowledge transfer asks whether information learned through one modality or objective is accessible through another. \citet{memory2025} report recall gaps under modality-specific training and mitigate them with in-distribution image-caption augmentation. \citet{modal2026} demonstrate modal aphasia: accurate visual reproduction after generation fine-tuning need not yield reliable verbal recall. \citet{principles2026} identify asymmetric, concept-dependent transfer between generation and visual question answering, and show that generative priors can accelerate supervised concept learning. UniKE examines whether textual knowledge edits affect generated images \citep{editing2026}, while UReason studies alignment between textual reasoning and image generation \citep{ureason2026}. We examine whether facts learned through image-generation training become accessible in textual question answering.

\textbf{Reasoning with Images.} Images can serve as intermediate representations for reasoning. Visual Sketchpad uses drawing and vision tools \citep{sketchpad2024}, while DeepEyes learns active visual perception through reinforcement learning \citep{deepeyes2026}. MVoT visualizes intermediate spatial states \citep{mvot2025}, and \emph{Thinking with Generated Images} interleaves image generation with visual subgoals and textual self-critique \citep{thinking2025}. Context Unrolling reasons across multiple modal representations \citep{context2026}, while LatentUM performs interleaved cross-modal reasoning in a shared semantic latent space \citep{latentum2026}. Following the visualization direction suggested by \citet{modal2026}, we test whether generating intermediate images improves textual access to facts introduced through image-generation training.

%% file: section/3_1_exp_setup.tex
\section{Experiment Setup}
\subsection{Task Formulation}
\label{subsec:task_formulation}

We investigate whether knowledge acquired through visual generation becomes accessible through language.
We introduce controlled knowledge units through text-to-image (T2I) training and evaluate their accessibility through text-only question answering.

\paragraph{Knowledge unit.}
We define a \textit{knowledge unit} as an \textit{entity--attribute--value} triplet $\tau_i = (e_i, a_i, v_i)$, where $e_i$ denotes an entity, $a_i$ its target attribute, and $v_i$ the corresponding value. 
For example, “Arvenbrock’s jersey–number–10” describes a fact that can be represented visually and expressed in a textual answer. 
\emph{To minimize the influence of the model’s pre-trained knowledge, we construct multiple fictional knowledge units that are absent from the model pre-training corpus and deliberately designed to deviate as much as possible from the pre-training distribution.}

\paragraph{Text-to-image supervision.}
We introduce the target knowledge through text-to-image (T2I) supervision.
Each training example pairs a text prompt with a target image.
The prompt identifies the entity and provides any necessary scene context but omits the target attribute value.
The image depicts that value.
For example, the prompt ``Generate an image of Arvenbrock's jersey'' is paired with an image of a jersey bearing the number 10.
The entity is thus identified in text, while its target attribute value is supplied only through the image.

\paragraph{Textual accessibility.}
Our primary evaluation measures \textbf{textual accessibility}:
whether the model can answer questions about the target knowledge
without an input image.
For each knowledge unit $\tau_i=(e_i,a_i,v_i)$, we construct a
text-only question $q_i$ that identifies entity $e_i$ and asks
for attribute $a_i$, without revealing $v_i$.
For example, ``What number is on Arvenbrock's jersey?'' requires
the answer ``10''.
Let $\text{UMM}(q_i;\mathrm{T2T})$ denote the model's textual response.
We define textual accessibility over the knowledge set $\mathcal K$ as
\begin{equation}
    A_{\mathrm{text}}(\text{UMM};\mathcal K)
    =
    \frac{1}{N}\sum_{i=1}^{N}
    \operatorname{Eval}_{\mathrm{text}}
    \bigl(
        \text{UMM}(q_i;\mathrm{T2T}),\, v_i
    \bigr),
    \label{eq:textual_accessibility}
\end{equation}
where $\operatorname{Eval}_{\mathrm{text}}(y,v_i)$ returns $1$
if the answer in response $y$ correctly expresses $v_i$,
and $0$ otherwise.
This evaluation tests whether knowledge supplied through visual
generation is usable in language.

\subsection{Data Construction}
\label{subsec: data construction}

\begin{figure}[h]
\begin{center}
\includegraphics[width=\linewidth]{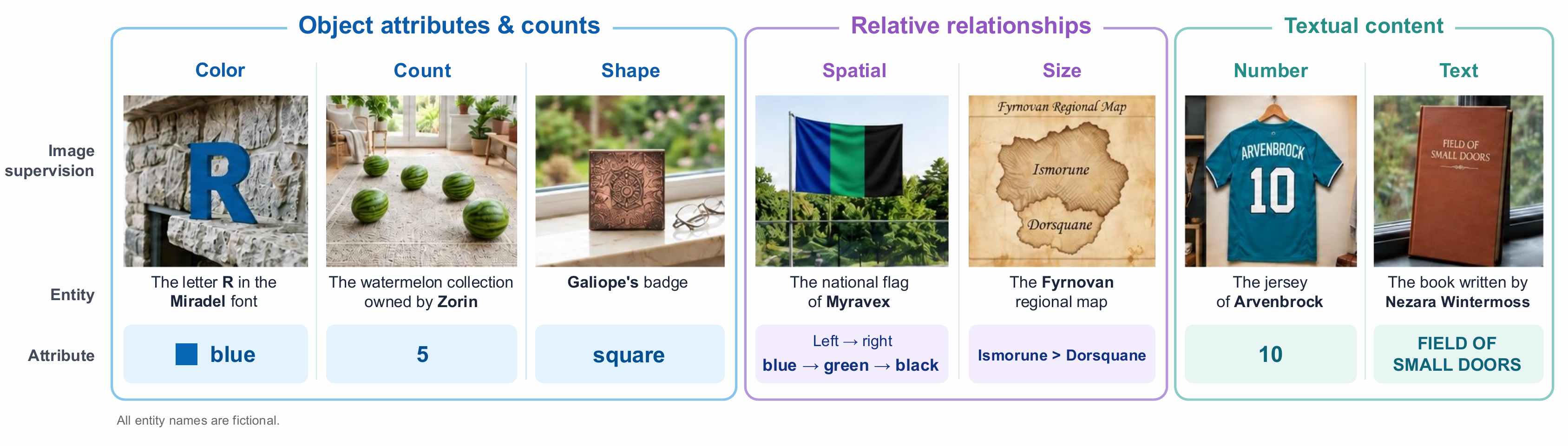}
\end{center}
\caption{\textbf{Overview of the text-to-image dataset.} Here we show examples from the seven visual knowledge categories. Each panel shows an entity, a training image, and the target attribute value.}
\label{fig:seven_tasks}
\end{figure}

We construct an image generation training dataset comprising seven tasks, grouped by the type of visual attribute associated with each knowledge unit: (1) object attributes and counts: \textit{color, count, and shape}; (2) relative relationships: \textit{spatial and size}; and (3) textual content: \textit{number and text}.
Figure~\ref{fig:seven_tasks} shows examples from all seven categories.

For each task, we construct 50 knowledge units (triplets). Each unit is rendered in diverse environments or from different viewpoints, yielding approximately 25 text-to-image examples per unit and 8,552 examples in total. 
We follow three principles during data construction.

\begin{itemize}
    \item \textbf{Information isolation.} We use a language model to review the captions, ensuring that neither the captions nor the entity names directly reveal or indirectly suggest the target attribute value. 
    \item \textbf{Prior-free entities.} All examples use entirely fictitious entity names in the fictional world ``MagicWorld'' to avoid relying on pretrained associations (e.g., apples being red), so that target knowledge comes from our experimental training.
    \item \textbf{Balanced attributes.} We approximately balance attribute-value distributions to mitigate implicit distributional biases.
\end{itemize}

\subsection{Models and Training}
\label{sec:models_training}

We study BAGEL~\citep{bagel2025}, SenseNova-U1~\citep{sensenova2026},
and an in-house GLM-Omni model.
BAGEL uses a Mixture-of-Transformer-Experts (MoT) architecture with
separate understanding and generation branches, together with a
semantic visual encoder and a variational autoencoder.
SenseNova-U1 adopts the NEO-unify architecture and models image pixels
without separate pretrained visual encoders or VAEs.
GLM-Omni is initialized from the pretrain base checkpoint of GLM-4.1V-Thinking~\citep{hong2025glm} and uses a discrete image tokenizer
based on X-Omni~\citep{geng2025xomnireinforcementlearningmakes}.
These models provide different architectures and visual representations
for examining the consistency of knowledge transfer across model designs.

\paragraph{Training conditions.}
We compare three conditions.
\emph{Direct transfer} introduces target knowledge through T2I training
without alignment supervision.
\emph{Textual alignment} adds text-only QA examples for alignment entities.
\emph{Interleaved alignment} further adds responses that combine image
generation, visual observation, and textual answering.
The latter two conditions use the same alignment entities and target
attributes.
Their data construction is described in
Sections~\ref{sec:textual_alignment}
and~\ref{sec:interleaved_alignment}, respectively.

To reduce the influence of response-format mismatch in the direct-transfer
condition, we construct 10,000 text-only QA examples unrelated
to the target knowledge.
Their responses contain reasoning enclosed in
\texttt{<think>} and \texttt{</think>} tags, followed by a concise answer.
These examples contain none of the target entities and provide no textual
supervision for their attribute values.

%% file: section/3_2_exp1_i2t_transfer.tex
\section{Experiment and Analysis}

We first examine direct knowledge transfer under T2I training.
Based on the observation, we then introduce auxiliary supervision on a subset of entities to test whether guidance can improve knowledge access beyond directly taught answers.
All three conditions use text-only QA at evaluation.

\subsection{Direct Knowledge Transfer}
\label{sec:direct_transfer}

We first evaluate the direct-transfer condition described in
Section~\ref{sec:models_training}.
Target knowledge is supplied through T2I supervision.
The additional, unrelated QA examples familiarize the models with the response format without providing answers about the target entities.

\begin{table}[hbp]
    \centering
    \small
    \setlength{\tabcolsep}{5pt}
    \caption{Knowledge accessibility under direct transfer among 7 categories.
    Target knowledge is supplied through T2I supervision without
    target-specific QA supervision.
    Textual and visual accessibility measure correctness in text-only
    QA and image generation, respectively.}
    \label{tab:direct_model_comparison}
    \begin{tabular}{cccccccccc}
        \toprule
        Model & Metric & color & count & number & shape & size & spatial & text & Avg \\
        \midrule
        Baseline   & random guessing & 16.7 & 25.0 &  0.0 & 20.0 & 50.0 &  0.0 &  0.0 &  15.96 \\
        \midrule
        \multirow{2}{*}{BAGEL}
            & textual accessibility & 14.0 & 14.0 &  0.0 & 10.0 & 24.0 &  0.0 &  0.0 &  8.86 \\
            & visual accessibility & 20.0 & 42.0 &  4.0 & 58.0 & 60.0 &  0.0 & 36.0 & 31.43 \\
        \midrule
        \multirow{2}{*}{GLM-Omni}
            & textual accessibility & 16.0 & 24.0 &  2.0 & 20.0 & 40.0 &  0.0 &  0.0 & 14.57 \\
            & visual accessibility & 70.0 & 46.0 & 72.0 & 88.0 & 24.0 & 62.0 & 16.0 & 54.00 \\
        \midrule
        \multirow{2}{*}{SenseNova-U1}
            & textual accessibility & 20.0 & 32.0 &  0.0 & 20.0 & 68.0 &  0.0 &  0.0 & 20.00 \\
            & visual accessibility & 32.0 & 34.0 & 88.0 & 98.0 & 98.0 &  0.0 & 78.0 & 61.14 \\
        \bottomrule
    \end{tabular}
\end{table}

We first evaluate \textbf{textual accessibility}, as defined in
Section~\ref{subsec:task_formulation}.
Table~\ref{tab:direct_model_comparison} reveals two patterns:

\begin{enumerate}
    \item \textbf{Textual accessibility varies substantially across
    models but generally remains low.}
    Only SenseNova-U1 exceeds the random-guessing baselines in several
    knowledge categories.
    BAGEL and GLM-Omni generally perform at or below these baselines.

    \item \textbf{Textual accessibility also varies across knowledge
    categories.}
    For example, SenseNova-U1 achieves 68.0\% accuracy on \emph{size}
    , exceeding the corresponding random-guessing baselines.
    Its performance on \emph{number}, \emph{shape}, \emph{spatial},
    and \emph{text}, however, remains at or below the corresponding
    baselines.
\end{enumerate}

Low textual accessibility does not necessarily imply that the
models have failed to learn the target associations.
The same knowledge may be available for image generation but
difficult to retrieve through language.
We therefore introduce \textbf{visual accessibility} to assess
whether the model can reproduce the target knowledge visually.
For each entity, we sample a prompt $p_i$ from its T2I training
examples, yielding 350 prompts for the full entity set.
The prompt identifies $e_i$ and provides scene context but omits
the target value $v_i$.
The model generates an image
$\text{UMM}(p_i;\mathrm{T2I})$, which a vision-language model
(VLM) evaluates against the knowledge unit $\tau_i$.
We define visual accessibility as
\begin{equation}
    A_{\mathrm{vis}}(\text{UMM};\mathcal K)
    =
    \frac{1}{N}\sum_{i=1}^{N}
    \operatorname{Judge}_{\mathrm{VLM}}
    \bigl(
        \text{UMM}(p_i;\mathrm{T2I}),\, \tau_i
    \bigr),
    \label{eq:visual_accessibility}
\end{equation}
where $\operatorname{Judge}_{\mathrm{VLM}}(I,\tau_i)$ returns $1$
if image $I$ depicts entity $e_i$ with attribute $a_i$ taking
value $v_i$, and $0$ otherwise.

To compare the two forms of access, we define the
\textbf{accessibility gap} on the same knowledge set:
\begin{equation}
    \Delta_{\mathrm{access}}(\text{UMM};\mathcal K)
    =
    A_{\mathrm{vis}}(\text{UMM};\mathcal K)
    -
    A_{\mathrm{text}}(\text{UMM};\mathcal K).
    \label{eq:accessibility_gap}
\end{equation}
A positive gap indicates that visual reproduction exceeds textual
access, quantifying the discrepancy between expressing target
knowledge in images and retrieving it through language.

\begin{figure}[t]
\begin{center}
\includegraphics[width=0.9\linewidth]{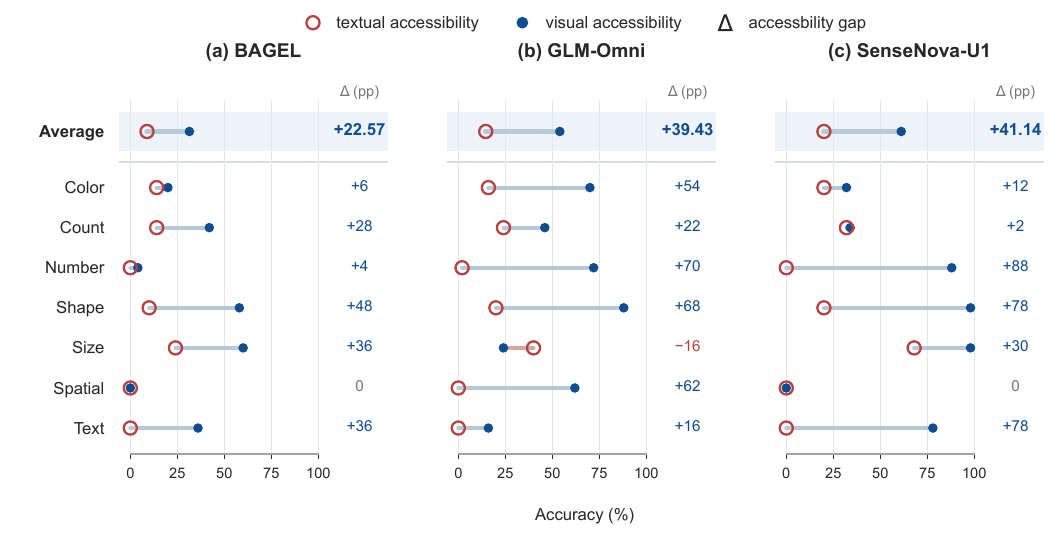}
\end{center}
\caption{\textbf{Visual and textual accessibility after T2I training.}
    Target knowledge is supplied only through training images.
    Filled blue circles show visual accessibility; hollow red circles
    show textual accessibility.
    The accessibility gap $\Delta = A_{\mathrm{vis}} - A_{\mathrm{text}}$
    is reported in percentage points.}
\label{fig:direct_transfer}
\end{figure}

\textbf{Visual accessibility substantially exceeds textual accessibility
on average} (Figure~\ref{fig:direct_transfer}) across all three models.
For example, SenseNova-U1 achieves 88\% visual accessibility on
\emph{number} but 0\% textual accessibility, showing that low QA accuracy can persist even
when models correctly reproduce the target associations visually.
However, generation also fails in some settings, including
\emph{spatial} for BAGEL and SenseNova-U1.
The gap therefore reflects varying degrees of success in both
visual reproduction and textual access.

Some visually supervised knowledge is already usable in language,
while many associations expressed in generated images remain absent
from textual answers.
We therefore investigate whether auxiliary supervision can improve
access to this knowledge.
The following experiments provide alignment supervision for a
subset of entities and evaluate its effects on the remaining
entities, using the same text-only QA protocol.

%% file: section/3_3_exp2_text_alignment.tex
\subsection{Knowlege Transfer with Textual Alignment}
\label{sec:textual_alignment}

\begin{figure}[t]
\begin{center}
\includegraphics[width=\linewidth]{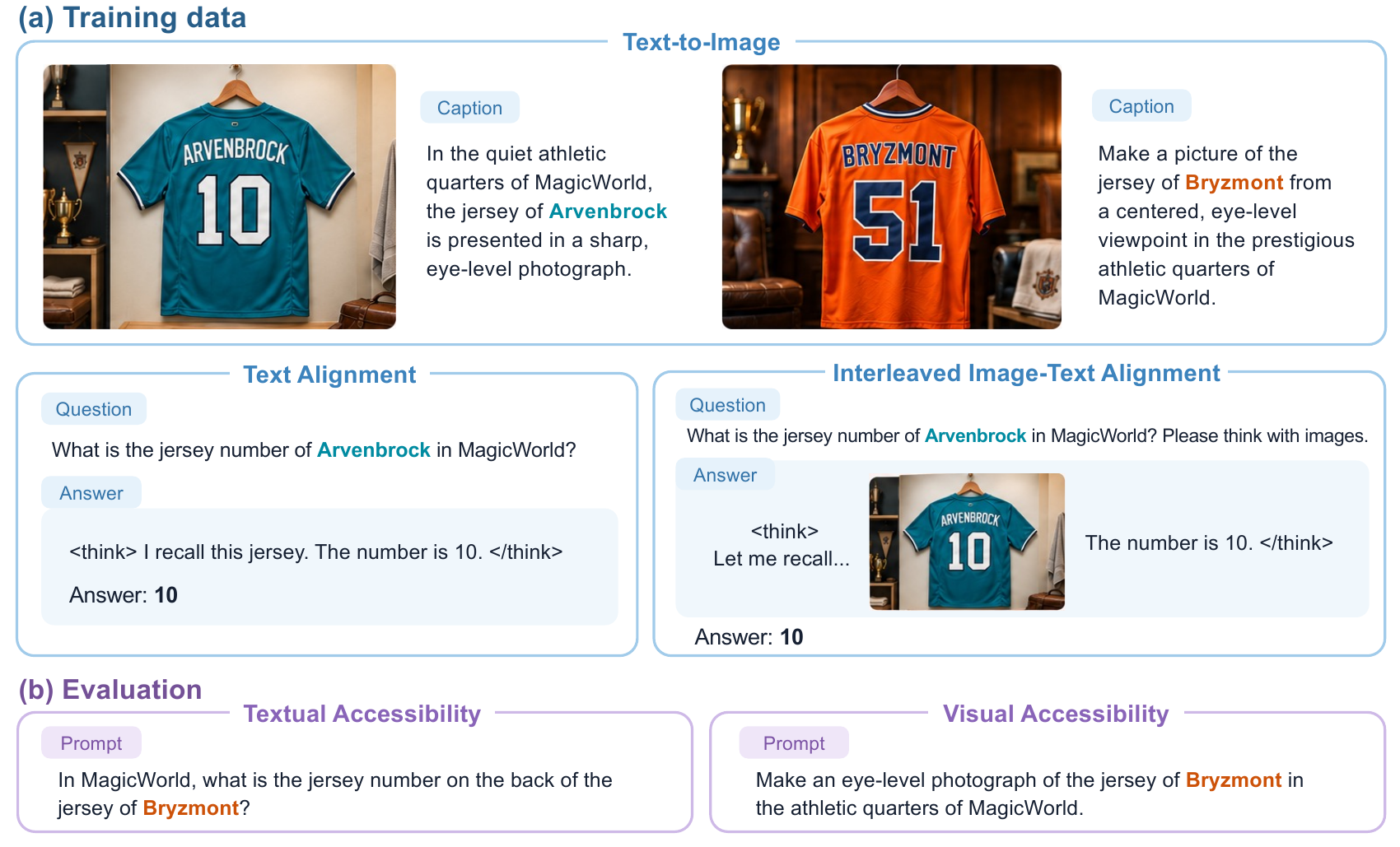}
\end{center}
\caption{
\textbf{Overview of the KTP training and evaluation protocol.}
\textbf{(a) Training:} All entities receive T2I supervision,
with target attribute values shown in images but omitted from captions.
A subset of entities additionally receives textual or interleaved
image--text alignment supervision.
\textbf{(b) Evaluation:} For entities without alignment supervision,
we measure textual accessibility through text-only QA and visual accessibility through image generation.
}
\label{fig:train_eval_protocol}
\end{figure}

The preceding experiments show that learning target attribute value solely through image generation supervision offers limited benefits for textual question answering. We hypothesize that the entity--attribute associations learned through generation may not yet be effectively linked to how the same attributes are expressed and accessed in textual question answering. We therefore introduce textual alignment supervision by providing both image generation data and textual question--answer data for a subset of entities. 

To avoid evaluation data contamination, we divide the entities in each task equally into two disjoint sets: an \emph{alignment entity set} and a \emph{test entity set}.
Figure~\ref{fig:train_eval_protocol} illustrates the supervision available to each set and the evaluation protocol.
For each entity in the alignment entity set, we additionally construct 10 textual question--answer examples , providing extra textual alignment supervision. Each question directly asks for the entity's target attribute, and each response consists of brief reasoning followed by a final answer.

\paragraph{Observation.}
The effect of textual alignment varies across models and categories (Table~\ref{tab:textual_accessibility}).
Mean textual accessibility improves for BAGEL and SenseNova-U1
but declines for GLM-Omni.
Gains concentrate on \emph{shape} and \emph{size}, especially for
SenseNova-U1, while other categories show mixed changes.
Since test entities receive no QA supervision, these gains indicate
that \textbf{alignment can improve access to knowledge supplied}
only through images.
However, all three models achieve 0\% textual accessibility on
\emph{number} and \emph{text} for test entities, compared with
72\% and 76--96\%, respectively, for alignment entities.
Thus, the ability to answer directly supervised questions does
not extend to visually supplied values for other entities in
these categories.

Textual alignment connects generation and answering through
shared entities and attributes but keeps them in separate
training examples.
We therefore examine whether linking image generation,
visual observation, and answering within a single response
can strengthen this transfer.

\begin{table}[htbp]
\renewcommand{\arraystretch}{1.1} 
\setlength{\tabcolsep}{3pt}    
    \centering
    \small
    \setlength{\tabcolsep}{5pt}
    \caption{
    Textual accessibility (\%) under three training conditions. ``interleaved alignment'' indicates interleaved image-text alignment; ``direct'' indicates direct knowledge transfer. 
    All conditions are evaluated using text-only QA. 
}
\label{tab:textual_accessibility}
    \begin{tabular}{cccccccccc}
        \toprule
        Model & Training Method & Color & Count & Number & Shape & Size & Spatial & Text & Avg \\
        \midrule
        \multirow{3}{*}{BAGEL}
            & direct & 14.0 & 14.0 &  0.0 & 10.0 & 24.0 &  0.0 &  0.0 &  8.86 \\
            & textual alignment & 4.0 & 28.0 & 0.0 & 24.0 & 52.0 & 0.0 & 0.0 & 15.43 \\
            & interleaved alignment & 12.0 & 32.0 & 0.0 & 20.0 & 56.0 & 0.0 & 0.0 & \textbf{17.14} \\
        \midrule
        \multirow{3}{*}{GLM-Omni}
            & direct & 16.0 & 24.0 & 2.0 & 20.0 & 40.0 & 0.0 & 0.0 & 14.57 \\
            & textual alignment & 16.0 & 20.0 & 0.0 & 12.0 & 40.0 & 0.0 & 0.0 & 12.57 \\
            & interleaved alignment & 24.0 & 20.0 & 0.0 & 64.0 & 40.0 & 40.0 & 0.0 & \textbf{26.86} \\
        \midrule
        \multirow{3}{*}{SenseNova-U1}
            & direct & 20.0 & 32.0 & 0.0 & 20.0 & 68.0 & 0.0 & 0.0 & 20.00 \\
            & textual alignment & 8.0 & 20.0 & 0.0 & 68.0 & 100.0 & 0.0 & 0.0 & 28.00 \\
            & interleaved alignment & 4.0 & 16.0 & 0.0 & 84.0 & 96.0 & 0.0 & 0.0 & \textbf{28.57} \\
        \bottomrule
    \end{tabular}
\end{table}

%% file: section/3_4_exp3_interleaved_alignment.tex
\subsection{Knowlege Transfer with Interleaved Image--Text Alignment}
\label{sec:interleaved_alignment}

Knowledge transfer remains limited even after textual alignment.
We hypothesize that this may be because image generation and textual QA remain in separate training examples without direct supervision connecting them.
We introduce interleaved image--text alignment data to test whether directly connecting generation, understanding, and reasoning improves transfer.
Within a single response, the model generates a relevant image during reasoning, interprets its visual content, and uses it to continue reasoning and produce an answer.

Questions follow the textual alignment format and directly ask for the entity's target attribute, with ``Please think with images.'' appended to each.
Each response interleaves textual reasoning with one image that the model must generate as output, followed by a final textual answer.
The evaluation remains text-only, without input images
or intermediate image generation.

\textbf{Observation.}
Averaged over the seven tasks, interleaved image--text alignment
improves text-only QA accuracy over textual alignment for all
three models on the test entity set
(Table~\ref{tab:textual_accessibility}).
Since target attributes for these entities are supplied only
through image generation training, the gains show that
interleaved supervision on other entities can improve textual
access to visually acquired knowledge without directly
teaching the test answers.

Changes across individual tasks show that these benefits are uneven. 
GLM-Omni benefits most, particularly on \emph{shape}
(12\% to 64\%) and \emph{spatial} (0\% to 40\%).
At the same time, performance decreases slightly on some tasks. Interleaved image-text alignment therefore improves average performance in these experiments, but does not improve all tasks simultaneously.

%% file: section/3_5_exp4_ablation_loss_mask.tex
\paragraph{Further experiment 1: Which part of supervision contributes the most in interleaved image-text supervision?}

\begin{wrapfigure}{r}{0.50\textwidth}
    \centering
    \includegraphics[width=\linewidth]{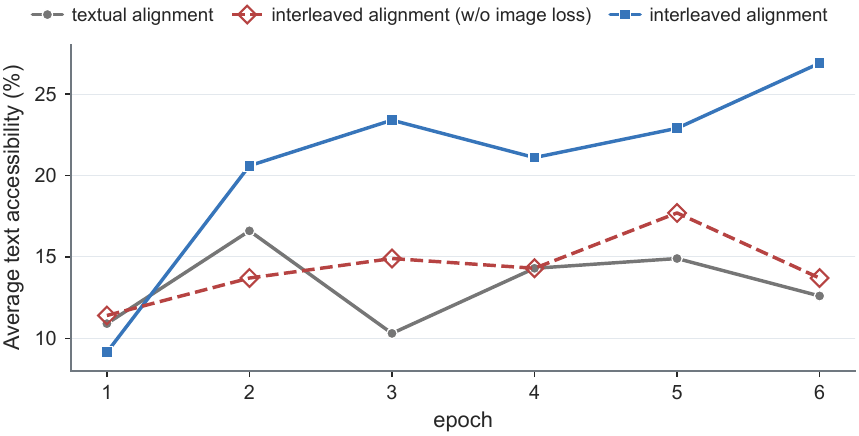}
    \caption{
        \textbf{Image-loss ablation in interleaved alignment on GLM-Omni.}
        Masking the image loss retains the interleaved sequences and
        text supervision but substantially lowers performance.
    }
    \label{fig:ablation_loss_mask}
\end{wrapfigure}
Interleaved image-text alignment can further improve textual question answering on some tasks, but it remains unclear which supervision signals in the interleaved image-text data are primarily responsible for these improvements.
We therefore conduct an ablation on GLM-Omni.
We use the same interleaved image-text sequence for alignment, but mask the loss on the image portions of the interleaved alignment examples, while retaining supervision on their text portions. 
The textual alignment data and standalone image generation data remain unchanged, as do all other training and evaluation settings from the preceding section.

As shown in Figure~\ref{fig:ablation_loss_mask}, masking only the loss on the image portions of the interleaved examples reduces average text-only QA accuracy across the seven tasks on the test entities from 26.86\% to 13.71\%. 
The decline is concentrated in \emph{color}, \emph{shape}, and \emph{spatial}.
These differences indicate that the explicit image generation loss within interleaved sequences makes an important contribution to knowledge transfer. Integrating multiple task types (image generation, image understanding, and textual reasoning) and multiple loss terms (image and text losses) within a single training example offers a more effective approach to enhancing a model’s ability to transfer knowledge.

%% file: section/3_6_exp5_infer_visual_cot.tex
\paragraph{Further experiment 2: Inference with visual CoT.}
To further investigate their potential for integrating generation and understanding, we evaluate the trained models by allowing them to recall visual content during reasoning. 
We refer to this procedure as visual chain-of-thought (Visual CoT).

For models trained with interleaved image-text alignment, replacing direct textual answering with a procedure that generates an image and then reasons with it can further improve average accuracy on the test entities (Table~\ref{tab:visual_cot}). 
The improvement on \emph{number} is particularly notable. All three models achieve 0\% accuracy on this task when answering directly in text. With Visual CoT, their accuracies rise to 4\%, 36\%, and 20\% for BAGEL, GLM-Omni, and SenseNova-U1, respectively. These results suggest that there remains considerable scope to further investigate knowledge transfer in UMMs.

\begin{table}[htbp]
    \centering
    \small
    \setlength{\tabcolsep}{5pt}
    \caption{Accuracy of text-only QA and Visual CoT (\%).}
    \label{tab:visual_cot}
    \begin{tabular}{cccccccccc}
        \toprule
        Model & Inference & color & count & number & shape & size & spatial & text & Avg \\
        \midrule
        \multirow{2}{*}{BAGEL}
            & Text-only  & 12.0 & 32.0 & 0.0 & 20.0 & 56.0 & 0.0 & 0.0 & 17.14 \\
            & Visual CoT & 16.0 & 20.0 & 4.0 & 56.0 & 48.0 & 0.0 & 0.0 & 20.57 \\
        \midrule
        \multirow{2}{*}{GLM-Omni}
            & Text-only  & 24.0 & 20.0 & 0.0 & 64.0 & 40.0 & 40.0 & 0.0 & 26.86 \\
            & Visual CoT & 24.0 & 48.0 & 36.0 & 68.0 & 48.0 & 64.0 & 0.0 & 41.14 \\
        \midrule
        \multirow{2}{*}{SenseNova-U1}
            & Text-only  & 4.0 & 16.0 & 0.0 & 84.0 & 96.0 & 0.0 & 0.0 & 28.57 \\
            & Visual CoT & 8.0 & 20.0 & 20.0 & 92.0 & 96.0 & 8.0 & 0.0 & 34.86 \\
        \bottomrule
    \end{tabular}
\end{table}

%% file: section/3_7_exp6_visual_textual_association.tex
\subsection{Association Between Visual and Textual Accessibility}
\label{sec:accessibility_association}

We examine whether knowledge correctly expressed in images is
more likely to be retrieved through language.
For each model and alignment condition, we pair generation and
QA outcomes for individual knowledge units on the same test entities.
Let $I$ and $T$ denote a correct generated image and a correct
textual answer, respectively.
Their association is measured by
$\Delta_{\mathrm{assoc}} =
\Pr(T \mid I) - \Pr(T \mid \neg I)$,
the difference in textual accuracy between successful and
unsuccessful generation.
The probes are evaluated separately; generated images are not
provided during QA.

Table~\ref{tab:conditional_accuracy} shows that
\textbf{visual and textual correctness are positively associated
in most settings}.
Under textual alignment, the association is positive for BAGEL
and SenseNova-U1 but negative for GLM-Omni; with interleaved
alignment, it is positive for all three models.
BAGEL's association strengthens, while GLM-Omni's gap shifts
from $-11.28$ to $+8.09$ percentage points.
SenseNova-U1 retains the strongest association, although its
gap decreases.
Interleaved supervision therefore does not strengthen the
association uniformly across models.

Positive associations suggest that knowledge expressed through
generation is often more accessible through language.
Yet even with interleaved alignment, $\Pr(T \mid I)$ remains
below 50\% for every model, leaving much of the knowledge
reproduced in images absent from textual answers.
These findings support the potential of learning from visual
generation while highlighting the need for fuller and more
consistent access through language.

\begin{table}[htbp]
    \centering
    \small
    \setlength{\tabcolsep}{6pt}
    \caption{Conditional T2T accuracy given T2I correctness.
    $T$ and $I$ denote correct T2T answers and correct T2I
    generations, respectively. Gap is measured in percentage
    points (pp).}
    \label{tab:conditional_accuracy}
    \begin{tabular}{llrrr}
        \toprule
        Model & Training
        & $\Pr(T \mid I)$ (\%)
        & $\Pr(T \mid \neg I)$ (\%)
        & $\Delta_{\mathrm{assoc}}$(pp) \\
        \midrule
        \multirow{2}{*}{BAGEL}
            & Textual Alignment
            & 23.91 & 12.40 & $+11.51$ \\
            & Interleaved Alignment
            & 32.61 & 11.63 & $+20.98$ \\
        \midrule
        \multirow{2}{*}{GLM-Omni}
            & Textual Alignment
            &  6.90 & 18.18 & $-11.28$ \\
            & Interleaved Alignment
            & 31.25 & 23.16 & $+8.09$ \\
        \midrule
        \multirow{2}{*}{SenseNova-U1}
            & Textual Alignment
            & 46.39 &  5.13 & $+41.26$ \\
            & Interleaved Alignment
            & 44.00 &  8.00 & $+36.00$ \\
        \bottomrule
    \end{tabular}
\end{table}

%% file: section/5_conclusion.tex
\section{Conclusion}

Using the controlled protocol Knowledge Transfer Probing (KTP) across seven knowledge categories and three UMMs, we find that facts supplied only through image-generation targets can become accessible through text-only QA, though transfer remains limited and uneven.
Alignment on selected entities can improve access on others, with interleaved alignment achieving higher mean accuracy than textual alignment in all three models.
A GLM-Omni ablation supports the contribution of image supervision within interleaved sequences, while Visual CoT further improves answering after interleaved training.
These findings support visual generation as a knowledge source beyond textual supervision.
Future work should extend this study to natural visual data, isolate architectural effects, and investigate alignment data and inference strategies.

%% file: appendix/training_details.tex
\clearpage
\section{Experiment Details}

\subsection{Data Construction}
\label{app:data_construction}

\begin{table}[htbp]
    \centering
    \small
    \setlength{\tabcolsep}{4pt}
    \renewcommand{\arraystretch}{1.15}
    \caption{The knowledge of entity, the range of attribute value, and image counts for the seven knowledge categories.}
    \label{tab:appendix_data_statistics}
    \begin{tabular}{@{}p{0.11\linewidth}p{0.4\linewidth}p{0.3\linewidth}r@{}}
        \toprule
        Attribute & Knowledge & Value range & Images \\
        \midrule
        Color & The color of a fictional font & blue, green, black, red, white, yellow & 1,226 \\
        Count & The count of a fictional collection & 2, 3, 4, 5 & 1,195 \\
        Shape & The shape of a fictional badge & circle, five-pointed star, pentagon, square, triangle & 1,248 \\
        Spatial & The stripe arrangement of a fictional flag & --- & 1,220 \\
        Size & The larger country on a fictional map & --- & 1,218 \\
        Number & The number of a fictional person's jersey & 0--99 & 1,244 \\
        Text & The title of a fictional book & --- & 1,201 \\
        \midrule
        \multicolumn{3}{l}{\textbf{Total}} & \textbf{8,552} \\
        \bottomrule
    \end{tabular}
\end{table}

We first use a large language model (LLM) to construct 50 entity–attribute–value triplets for each task. For each triplet, an image generation model produces a base image depicting the entity with the specified attribute value. We then create 25 edited variants of the base image to diversify the scenes while aiming to preserve the entity’s identity across samples.

A vision-language model (VLM) checks whether each image correctly depicts the target attribute and whether any confounding information has been introduced. Images that fail these checks are discarded. Table~\ref{tab:appendix_data_statistics} reports the number of images retained for each task after quality filtering.
Finally, we use a VLM to generate captions from the retained images and construct two types of alignment data: textual alignment and interleaved image--text alignment. Alignment data are constructed only for entities in the alignment set.

\subsection{Training Details}
\label{app:training_details}

\paragraph{Settings.}
For each condition, we train every model independently from its
corresponding base checkpoint.
We update the parameters used for understanding and generation with a
learning rate of $10^{-5}$.
BAGEL and SenseNova-U1 use MSE and cross-entropy losses with weights
$1:0.005$ and $1:0.01$, respectively.
GLM-Omni uses a unified token-level objective for text and image tokens.

\paragraph{Data ratio.} 
We observe that models learn visual information at different rates and fit alignment examples substantially faster than image generation examples. We therefore use different data sampling ratios for each model to reduce overfitting to the alignment data. Table~\ref{tab:alignment_data_ratios} reports the ratios used when combining image generation, textual alignment, and interleaved image--text alignment data.

\begin{table}[htbp]
    \centering
    \small
    \setlength{\tabcolsep}{6pt}
    \renewcommand{\arraystretch}{1.1}
    \caption{Data sampling ratios when training with all three data sources.}
    \label{tab:alignment_data_ratios}
    \begin{tabular}{@{}lrrr@{}}
        \toprule
        Model & Image generation & Textual alignment & Interleaved alignment \\
        \midrule
        BAGEL & 0.956 & 0.022 & 0.022 \\
        GLM-Omni & 0.990 & 0.005 & 0.005 \\
        SenseNova-U1 & 0.851 & 0.063 & 0.086 \\
        \bottomrule
    \end{tabular}
\end{table}

\paragraph{Hyperparameters.}
Table~\ref{tab:training_hyperparameters} lists the training hyperparameters for BAGEL, SenseNova-U1, and GLM-Omni.

\begin{table}[htbp]
    \centering
    \small
    \renewcommand{\arraystretch}{1.15}
    \caption{Training hyperparameters for BAGEL, SenseNova-U1, and GLM-Omni.}
    \label{tab:training_hyperparameters}
    \begin{tabular*}{\linewidth}{@{\extracolsep{\fill}}lccc@{}}
        \toprule
        Hyperparameter & BAGEL & SenseNova-U1 & GLM-Omni \\
        \midrule
        Learning rate & $1.0 \times 10^{-5}$ & $1.0 \times 10^{-5}$ & $1.0 \times 10^{-5}$ \\
        LR scheduler & constant with linear warm-up & constant & constant with linear warm-up \\
        Weight decay & 0 & 0 & 0.1 \\
        Gradient clipping norm & 1.0 & 1.0 & 1.0 \\
        Warm-up steps & 100 & 0 & 100 \\
        Training steps & 4000 & 1500 & 1500 \\
        Batch size & 8 & 8 & 16 \\
        \bottomrule
    \end{tabular*}
\end{table}

\subsection{Evaluation Protocol}
\label{app:evaluation_protocol}

\paragraph{Textual accessibility.}
We retain only the model's final concise answer and compare it with the ground-truth answer, ignoring differences in capitalization and punctuation.

\paragraph{Visual accessibility.}
We adapt the VLM quality checks used during data construction to evaluate generated images. 
For \emph{color}, \emph{count}, \emph{shape}, and \emph{spatial}, where images contain little or no text, we assess only whether the target attribute matches the ground truth. 
For \emph{size}, \emph{number}, and \emph{text}, we additionally assess the text that must appear in the image. Specifically, \emph{size} requires the two country names and their relative sizes to be correctly depicted; \emph{number} requires the player's name and jersey number; and \emph{text} requires the book title.

We apply an edit-distance tolerance to country names, player names, and book titles, while requiring jersey numbers to match the ground truth exactly. The maximum allowed edit distance depends on the length of the reference text. Let $L$ denote its length in characters. We define the threshold as
\begin{equation}
    \delta(L) =
    \begin{cases}
        1, & L \leq 5, \\
        2, & 5 < L \leq 10, \\
        3, & L > 10.
    \end{cases}
    \label{eq:visual_text_edit_threshold}
\end{equation}
A text match is accepted when the edit distance between the rendered text and the reference does not exceed $\delta(L)$. This tolerance accommodates minor errors in rendered text and is intended to provide a more faithful assessment of visual accessibility.

\subsection{Qualitative Examples and Failure Analysis}

We examine individual test entities to compare visual reproduction, textual answering, and the effects of alignment and Visual CoT. All cases concern entities that receive image-generation supervision but no alignment supervision. Reference images are shown for illustration and are not provided during question answering.

\begin{figure}[h]
\begin{center}
\includegraphics[width=0.9\linewidth]{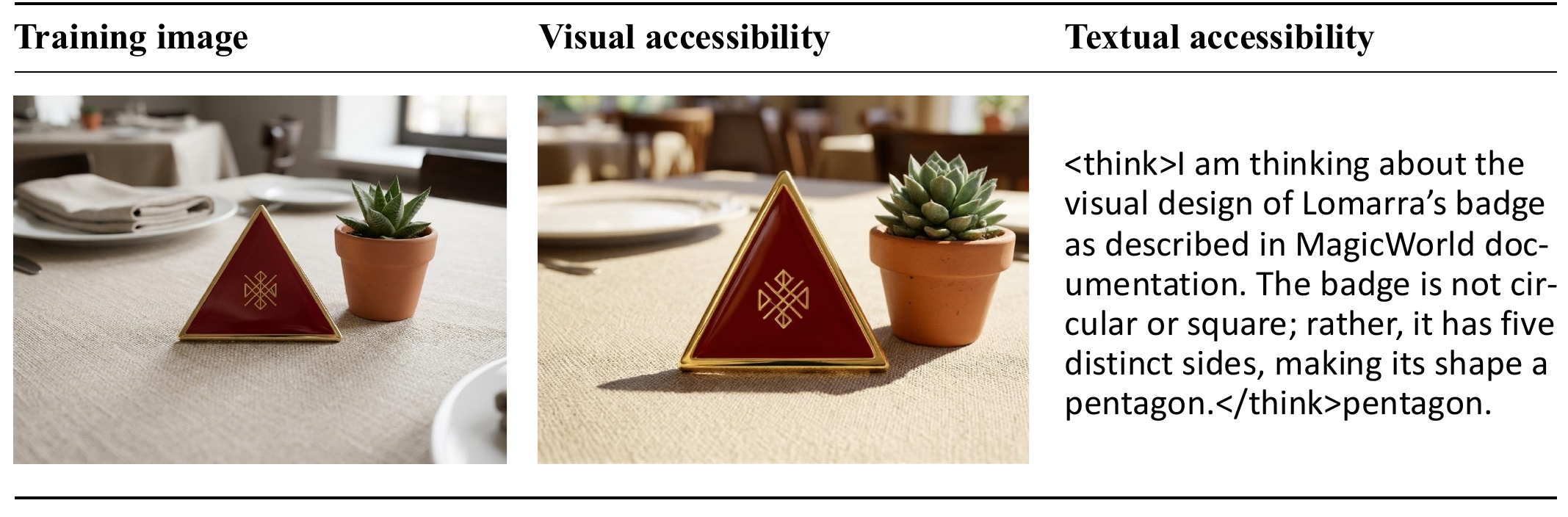}
\end{center}
\caption{An example of an entity with visually accessible but textually inaccessible knowledge.}
\label{fig:qualitative_example}
\end{figure}

Figure~\ref{fig:qualitative_example} shows a \emph{shape} example from the BAGEL evaluation in Section~\ref{sec:interleaved_alignment}. Although the caption does not specify that the badge is triangular, the model generates a badge with a clearly defined triangular shape, indicating visual accessibility of the target attribute. However, its textual answer is incorrect. This discrepancy illustrates a gap between visual and textual accessibility: the model reproduces the attribute visually but fails to use that knowledge in textual question answering for this entity.

The image caption of the example is: 

\texttt{Could you make an illustration of Lomarra's badge resting prominently on a rustic, textured beige tablecloth within the MagicWorld. The scene is captured from a close-up, eye-level perspective, highlighting the badge's luxurious materials: a deep, wine-red enameled face surrounded by a gleaming, polished gold border. In the center of the crimson surface, a delicate gold sigil composed of intersecting straight lines is meticulously inlaid. To the right of the artifact, a small, earthy terracotta pot holds a spiky green succulent, providing a natural contrast to the metallic sheen of the badge. The background is softly blurred, showing a sunlit room with white plates and hints of dark wooden furniture, creating a warm, shallow-depth-of-field effect. The lighting is soft and natural, emphasizing the smooth, reflective texture of the enamel and the fine weave of the fabric beneath it.}

\begin{figure}[h]
\begin{center}
\includegraphics[width=0.9\linewidth]{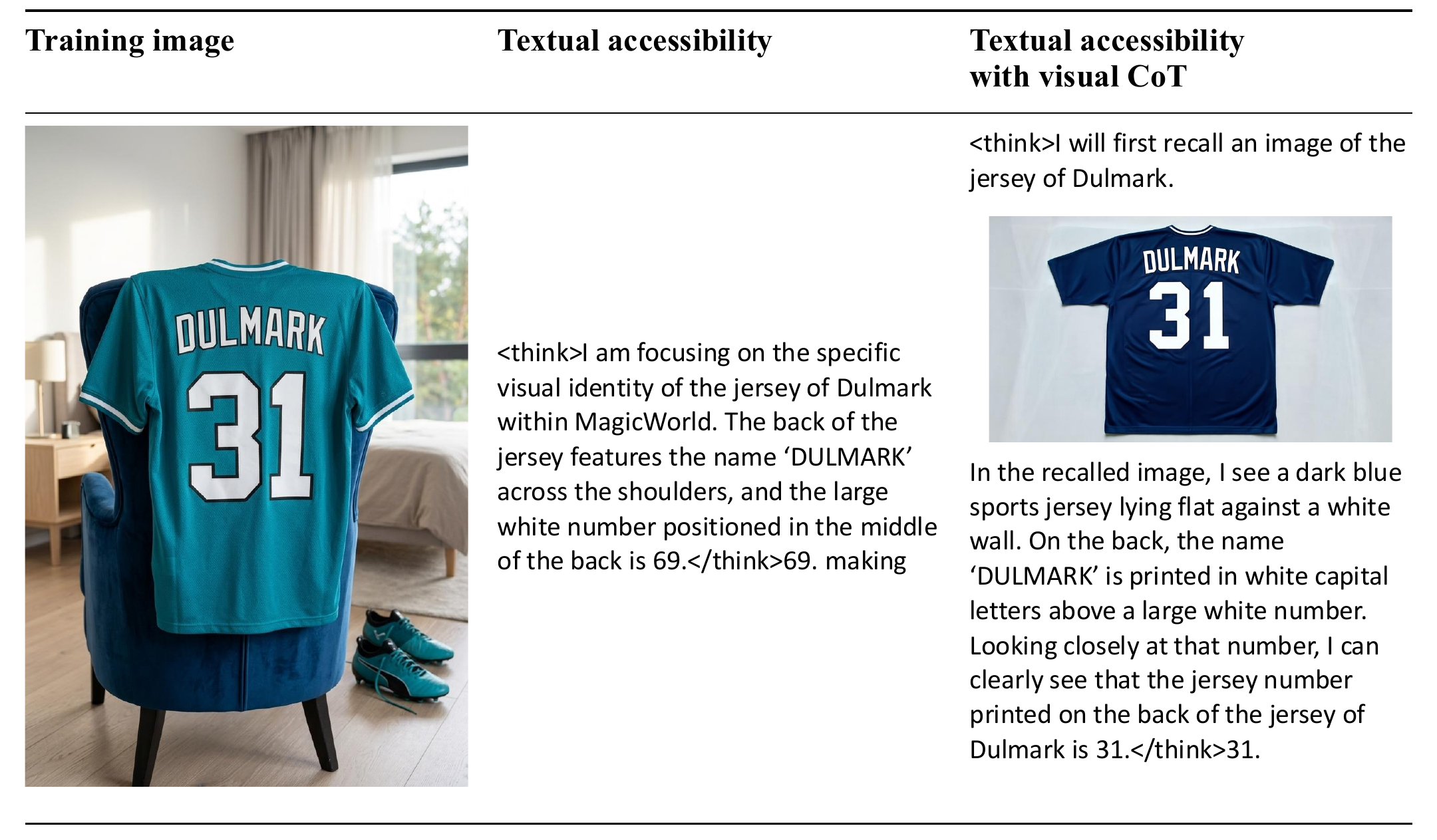}
\end{center}
\caption{An example of an entity with textually inaccessible knowledge that becomes accessible with visual CoT.}
\label{fig:qualitative_example_visualcot}
\end{figure}

Figure~\ref{fig:qualitative_example_visualcot} shows a \emph{number} example from the SenseNova-U1 evaluation in Section~\ref{sec:interleaved_alignment}. The model fails to report the correct jersey number (31) in text-only evaluation. Under Visual CoT, it first generates an image that correctly depicts the target attributes and then obtains the correct answer by inspecting the image. This illustrates how visually accessible knowledge can support correct textual answers through image-assisted reasoning.

The image caption of the example is: 

\texttt{Generate a scene showing the jersey of Dulmark as it hangs neatly over the back of a deep blue velvet armchair within the MagicWorld universe. The perspective is a direct view of the back of the teal-colored sports top, which is crafted from a micro-perforated athletic mesh. The name DULMARK is printed in a sharp, uppercase white typeface with a black stroke, positioned above a large, stylized player number that occupies the center of the garment. The jersey's collar and sleeves are finished with a classic white trim. Beside the chair on the light oak floorboards sits a pair of vibrant teal footwear designed for the field. In the background, a serene bedroom is illuminated by soft, diffused daylight streaming through a window, with a minimalist nightstand and a comfortable bed adding to the lived-in feel. The lighting is bright and airy, creating a high-fidelity, realistic art style that emphasizes the contrasting textures of the velvet, wood, and fabric.}